\documentclass[sigconf]{acmart}
\AtBeginDocument{%
  \providecommand\BibTeX{{%
    \normalfont B\kern-0.5em{\scshape i\kern-0.25em b}\kern-0.8em\TeX}}}
\usepackage{multirow}
\usepackage{multicol}

\usepackage{amssymb}
\usepackage{enumitem}
\usepackage{url}
\usepackage{balance} 
\setlist[itemize,1]{leftmargin=\dimexpr 26pt-2mm}

\copyrightyear{2024}
\acmYear{2024}
\setcopyright{acmlicensed}
\acmConference[KDD '24] {Proceedings of the 30th ACM SIGKDD Conference on Knowledge Discovery and Data Mining }{August 25--29, 2024}{Barcelona, Spain.}
\acmBooktitle{Proceedings of the 30th ACM SIGKDD Conference on Knowledge Discovery and Data Mining (KDD '24), August 25--29, 2024, Barcelona, Spain}
\acmISBN{979-8-4007-0490-1/24/08}
\acmDOI{10.1145/3637528.3671579}
\begin{document}

\title[DuMapNet: An End-to-End Vectorization System for City-Scale Lane-Level Map Generation]{DuMapNet: An End-to-End Vectorization System for \\ City-Scale Lane-Level Map Generation}

\author{Deguo Xia}
\authornote{These authors contributed equally to this work.}
\affiliation{%
  \institution{Tsinghua University}
  \state{Beijing}
  \country{China}}
\affiliation{%
  \institution{Baidu Inc.}
  \state{Beijing}
  \country{China}}
\orcid{0000-0003-3366-2230}
\email{xiadeguo@baidu.com}

\author{Weiming Zhang}
\authornotemark[1]
\orcid{0000-0003-2609-2807}
\affiliation{%
  \institution{Baidu Inc.}
  \state{Beijing}
  \country{China}}
\email{zhangweiming@baidu.com}

\author{Xiyan Liu}
\authornotemark[1]
\orcid{0000-0002-0102-9636}
\affiliation{%
  \institution{Baidu Inc.}
  \state{Beijing}
  \country{China}}
\email{liuxiyan@baidu.com}
  
\author{Wei Zhang}
\authornotemark[1]
\orcid{0000-0003-3900-1222}
\affiliation{%
  \institution{Baidu Inc.}
  \state{Beijing}
  \country{China}}
\email{zhangwei99@baidu.com}

\author{Chenting Gong}
\authornotemark[1]
\orcid{0009-0001-4445-9361}
\affiliation{%
  \institution{Baidu Inc.}
  \state{Beijing}
  \country{China}}
\email{gongchenting@baidu.com}
  
\author{Jizhou Huang}
\authornote{Corresponding authors.}
\orcid{0000-0003-1022-0309}
\affiliation{%
  \institution{Baidu Inc.}
  \state{Beijing}
  \country{China}}
\email{huangjizhou01@baidu.com}

\author{Mengmeng Yang}
\authornotemark[2]
\orcid{0000-0002-3294-6437}
\affiliation{%
  \institution{Tsinghua University}
  \state{Beijing}
  \country{China}}
\email{yangmm_qh@tsinghua.edu.cn}
  
\author{Diange Yang}
\orcid{0000-0002-0074-2448}
\affiliation{%
  \institution{Tsinghua University}
  \state{Beijing}
  \country{China}}
\email{ydg@mail.tsinghua.edu.cn}

\renewcommand{\shortauthors}{Deguo Xia et al.}

\begin{abstract}
Generating city-scale lane-level maps faces significant challenges due to the intricate urban environments, such as blurred or absent lane markings. Additionally, a standard lane-level map requires a comprehensive organization of lane groupings, encompassing lane direction, style, boundary, and topology, yet has not been thoroughly examined in prior research. These obstacles result in labor-intensive human annotation and high maintenance costs. This paper overcomes these limitations and presents an industrial-grade solution named DuMapNet that outputs standardized, vectorized map elements and their topology in an end-to-end paradigm. To this end, we propose a group-wise lane prediction (GLP) system that outputs vectorized results of lane groups by meticulously tailoring a transformer-based network. Meanwhile, to enhance generalization in challenging scenarios, such as road wear and occlusions, as well as to improve global consistency, a contextual prompts encoder (CPE) module is proposed, which leverages the predicted results of spatial neighborhoods as contextual information. Extensive experiments conducted on large-scale real-world datasets demonstrate the superiority and effectiveness of DuMapNet. Additionally, DuMapNet has already been deployed in production at Baidu Maps since June 2023, supporting lane-level map generation tasks for over $360$ cities while bringing a $95\%$ reduction in costs. This demonstrates that DuMapNet serves as a practical and cost-effective industrial solution for city-scale lane-level map generation.

\end{abstract}


\maketitle

\begin{figure}%
\includegraphics[width=1.0\linewidth,trim={0.07cm 0.13cm 0.07cm 0.0cm},clip]{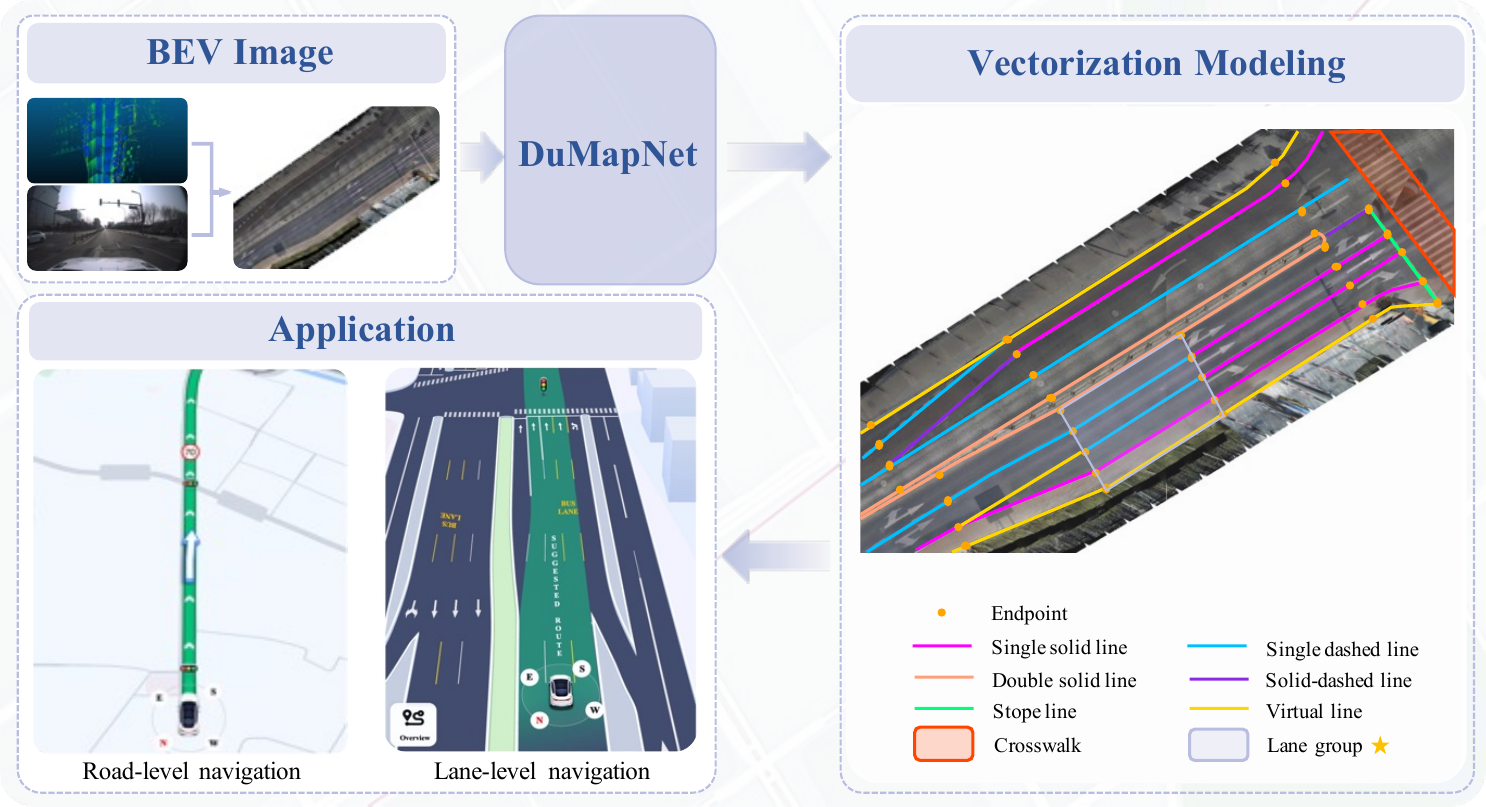}
\caption{
DuMapNet introduces a learning-based methodology for lane-level map vectorization. Our proposed method incorporates a scheme based on contextual prompts and components dedicated to group-wise lane prediction. With these advancements, DuMapNet achieves cost-effective generation of city-scale vectorized maps and significantly supports various applications, such as lane-level navigation in Baidu Maps.}
\label{fig:ad}
\end{figure}

\section{Introduction}
Lane-level map, as a crucial layer of the high-definition map, offers critical prior information for autonomous driving, facilitating beyond visual line of sight (BVLOS) perception, path planning, and decision-making with globally consistent road data. Specifically, lane-level map models the real world within decimeter level. This indicates that they not only depict the fundamental structure and layout of roads but also provide lane-level details, including lane line geometry, lane marking style, as well as connection topology, \textit{etc.} Building on this, lane-level navigation, a pioneering advancement in precise travel guidance, has been extensively deployed to assist public travel by providing detailed, high-precision map elements and suggested routes (see Figure~\ref{fig:ad}). Additionally, with its higher precision and advanced route planning capabilities, lane-level map will potentially benefit a wide range of geographical tasks, such as geo-object change detection~\cite{duarus2022}, traffic condition prediction~\cite{dutraffic2022}, estimated time of arrival prediction (ETA prediction, a.k.a., travel time estimation)~\cite{dueta2022, fang2020constgat, fang2021ssml}, and road extraction~\cite{duare2022} at Baidu Maps.

The task of lane-level map generation can be formulated as constructing and updating the core geographic elements that meet lane-level precision, given abundant road data. These geographic elements primarily include open-shape elements (\textit{e.g.}, lane lines, stop lines, \textit{etc.}) and closed-shape elements (\textit{e.g.}, crosswalks). Moreover, as an efficient and standardized management unit in high-definition map, lane group plays a crucial role in onboard navigation and autonomous driving. It can be defined as a set of one or more lanes on a road segment perpendicular to the direction of travel~\cite{NDS,krausz2022comparison} (see Figure~\ref{fig:ad}). Within a lane group, the number of lanes remains constant, and all lanes belong to the same road segment with the same direction of travel. Given this definition, the lane group emerges as an exceptionally convenient and efficient unit for driving guidance, markedly improving vehicle interaction with urban environments. Consequently, this paper introduces a learning-based solution that directly generates final standardized results in an end-to-end manner. This method effectively supersedes existing techniques that rely on manual post-processing to construct lane groups. 

Traditional map generation solutions are often costly and labor-intensive as they require trained experts to manually annotate geographic elements. To improve efficiency with less human effort, leveraging advancements in computer vision for map generation has become a viable approach. These algorithms can be roughly categorized into segmentation-based methods~\cite{tao2020hierarchical,li2022hdmapnet,peng2023bevsegformer}, lane detection-based methods~\cite{feng2022bezierlaneNet,liu2021condlanenet}, as well as vectorization-based methods~\cite{liao2022maptr,liu2023vectormapnet,zhang2023gemap}. Specifically, segmentation-based methods are suboptimal since they often require a series of post-processing strategies, such as thinning and fitting, to convert mask into vectorized map. Lane detection-based methods are usually limited in terms of extensibility and flexibility regarding map element types. While the vectorization-based solutions have achieved commendable results, they still exhibit limitations in prediction accuracy, post-processing logic and handling complex road scenarios, such as road wear and vehicle occlusion. Moreover, such onboard approaches are often constrained by computational power and local construction patterns, preventing them from meeting the precision and global consistency required for city-scale lane-level map.

To fully explore the paradigm of large-scale lane-level map generation, we propose an automatic industrial-grade offboard solution termed DuMapNet. Given a bird's-eye-view (BEV) image, DuMapNet can unify the modeling of polyline-style and polygon-style map elements as a set of points. To significantly improve the prediction results for difficult scenarios such as road wear, occlusions, and complex intersections, as well as the connections of vectorization results among frames, we propose the \textbf{contextual prompts encoder (CPE)} module. By using the spatial prediction results of the current BEV image's neighborhood as prompt information, CPE significantly enhances the geometric and category consistency of the prediction results in a larger receptive field. To avoid the error accumulation effect and weak generalization of traditional multi-stage map-making methods, and considering the requirements for standardized map construction, we design a \textbf{group-wise lane prediction (GLP)} to output the vectorized results of lane groups through mutual constraints between lane group polygons and lane lines, without the need for complex post-processing logic. Finally, to achieve an end-to-end large-scale map generation mode, we develop the \textbf{topology prediction} module, which predicts the lane line topological relationships between BEV images, enabling large-scale map correlation. Our key contributions to both the research and industrial communities are as follows:

\begin{itemize}
\item \textbf{Potential impact:} We introduce DuMapNet, an end-to-end vectorization modeling framework, as an industrial-grade solution for city-scale lane-level map generation. DuMapNet has been successfully deployed in production at Baidu Maps, supporting lane-level map generation for over $360$ cities and realizing a $95\%$ reduction in costs.
\item \textbf{Novelty:} DuMapNet represents a new paradigm for city-scale lane-level map generation task, achieving end-to-end predictions from bird's-eye-view (BEV) images to vectorized results that meet cartographic standards. The novelty lies in each stage, from the unified vectorization modeling, the group-wise lane
prediction system, the contextual prompts encoder, to the topology prediction module, making the lane-level map generation task highly automatic and cost-effective.
\item \textbf{Technical quality:} Extensive qualitative and quantitative experiments are performed on large-scale, real-world datasets collected from Baidu Maps, which demonstrate the superiority of DuMapNet. The successful deployment of DuMapNet at Baidu Maps further shows that it is a practical and robust solution for city-scale lane-level map generation.
\end{itemize}

\section{D\texorpdfstring{\MakeLowercase {u}}{u}M\texorpdfstring{\MakeLowercase {ap}}{ap}N\texorpdfstring{\MakeLowercase {et}}{et}} \label{section:dumapnet}

\begin{figure*}
\includegraphics[width=1.0\linewidth,trim={0.0cm 0.05cm 0.8cm 0.0cm},clip]{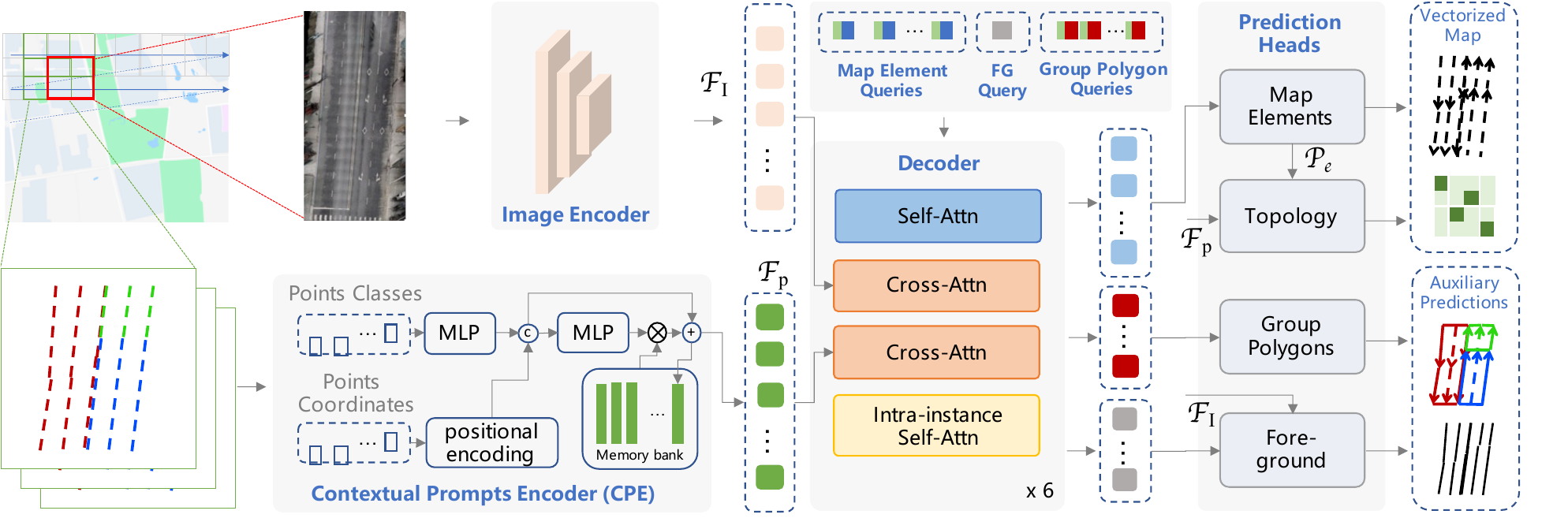}
\vspace{-5mm}
\caption{Overall architecture of DuMapNet. DuMapNet processes the entire city-scale land area using a sliding window approach. For each local area, an image encoder is utilized to extract image features from the BEV image. Meanwhile, we propose a novel Contextual Prompts Encoder (CPE) to encode the predictions of adjacent scanned areas. To achieve Group-wise Lane Prediction (GLP), we meticulously tailor key network components, including the query, decoder, and prediction heads. Consequently, the network is capable of generating a vectorized map, which encompasses vectorized elements and their topology. Additionally, two auxiliary predictions are generated: the use of group polygons aids in the organization of lane groups, while foreground segmentation enhances lane point localization. For detailed illustrations, please refer to Section \ref{section:dumapnet}.}
\vspace{-2mm}
\label{fig:architecture}
\end{figure*}

\subsection{Preliminaries}
The task of lane-level map generation from BEV image is defined as follows: given a BEV image $I$ collected from vehicle-mounted sensors as input, the network is supposed to predict the vectorized map elements $V$. Next, we will describe the data preparation and unified vectorization.

\textbf{Data Preparation. }
Different from most onboard methods that operate on BEV features using multi-view images as input, following~\cite{ammar2019geometric,kim2021scan}, our offboard approach is built upon the BEV image produced using multi-view images, point cloud data, and vehicle poses information. The advantages primarily lie in two aspects: first, the regional global information can be fully utilized, such as geometric smoothness constraints, semantic correlations, and global precision consistency; second, conducting multi-trip data collection can alleviate the inevitable challenges such as precision bias and dynamic occlusion. Instead of appealing to heavy labeling manpower, we solve the large-scale annotation problem using the Baidu Map Database in an automatic fashion. Specifically, given a BEV image $I$ with $H \times W$ resolution, with the spatial resolution $4cm \times 4cm$ for each pixel, it covers $H / 25$ meters by $W / 25$ meters region with a certain geographic coordinate range. First, we index the instance geometries, labels, and lane group IDs within that range from the database. Second, based on the lane group IDs, the instance geometries are organized into a list format at the granularity of lane groups and mapped to pixel coordinate system. Simultaneously, we compute the minimum bounding rectangle of all instances contained within each lane group to create the group polygon. Finally, based on the spatial relationship between BEV images, neighborhood IDs are added to the ground truth. For better understanding, we have released a demo for reference on GitHub at \url{https://github.com/XiyanLiu/DuMapNet}.

\textbf{Unified Vectorization. }
We define a unified vectorized representation for the core geographic elements in each local land area. Formally, given a BEV image $I$, we denote the corresponding lane-level vectorization as $V=\{G_{i}\}^{N_{g}}_{i=0}$, where ${N_{g}}$ denotes the number of the lane groups in the local land area. Each lane group $G_{i}$ is composed of a set of geographic elements $P_{j}$ and element style $C_{j} \in \mathbb{R}^{N_{cls}}$, where $C_{j}$ is a one-hot vector with $N_{cls}$ element styles in total. We thus denote a lane group as $G=\{P_{j}, C_{j}\}^{N_{l}}_{j=0}$, where ${N_{l}}$ is the element number in a lane group. Next, the point set of each element instance is denoted as $P=\{p_{k}\}^{N_{p}}_{k=0}$, where $N_{p}$ is the number of the points and $p_{k} \in \mathbb{R}^2$ denotes the coordinates of each point.

\subsection{Overall Architecture}
The city-scale lane-level map generation is inherently complex, requiring a comprehensive organization of lane groupings and existing methods only generate partial elements of lane groupings. Meanwhile, DuMapNet is the first end-to-end solution to achieve city-scale lane-level map generation, realizing practical and effective industrial gains.

Specifically, Figure~\ref{fig:architecture} illustrates the overall architecture of our proposed DuMapNet. To obtain a city-scale lane-level vectorized map, DuMapNet processes the entire land area using a sliding window approach following a zig-zag scan sequence. The model's inputs contain two parts: a BEV image and contextual prompts. Specifically, the BEV image is obtained by the aforementioned data preparation process, providing abundant appearance features of the local land area. We employ an image encoder that contains a backbone network and a Feature Pyramid Network (FPN) to extract BEV feature $F_{I} \in \mathbb{R}^{384\times 384}$ from the BEV image $I$. Meanwhile, to enable spatial coherent lane-level vectorized predictions over adjacent land areas, we propose to take the predicted vectorized map of adjacent scanned areas as the additional input of DuMapNet. We further tailor a contextual prompts encoder (CPE) to realize an effective encoding for the predictions of adjacent scanned areas. Moreover, we devise a query combination that contains a set of hierarchical queries for lane line prediction, one foreground segmentation query, and a set of queries for lane group polygon prediction. In the Decoder, the proposed query combination interacts with both the BEV feature and the contextual prompt embeddings derived from CPE. Finally, we construct multiple tasking prediction heads to facilitate various predictions, where the predictions include vectorized results for lane lines and the topology that indicates the connection of lane lines between different lane areas. Furthermore, we propose two auxiliary predictions, where the group polygons are adopted to facilitate lane group organization while the foreground segmentation is to help improve lane point localization. 

\subsection{Contextual Prompts Encoder (CPE)} 

Inspired by the recent success of prompt-based vision models~\cite{kirillov2023segment, jia2022visual, liu2023visual}, our proposed CPE adopts a simple yet effective architecture to encode both the geometric and semantic information of the vectorization results of adjacent land areas, providing contextual cues for the vectorization of the current land area during the sliding window operation. Formally, we define
\begin{equation}
\label{equ1}
F_p = \mathbb{CPE}(\{V_{s}\}_{s\in A})
\end{equation}
where $A$ denotes a set of adjacent land areas. $F_p \in R^{M_{g}\times M_{l} \times{N_{p}}\times256}$ denotes the prompt embeddings that interact with the features of intermediate layers in the decoder. Where $M_{g}, M_{l}, N_{p}$ denote the number of total predicted groups, the number of element instances in each group, and the number of points in each instance, respectively. 

The architecture of CPE is illustrated in Figure~\ref{fig:architecture}. Specifically, we adopt a shared MLP (Multi-Layer Percseptron) network to encode the predicted style type for each element to serve as the semantic encoding. For the geometric information, we perform a shared positional encoding sub-network to the coordinates of element points. This sub-network consists of sine and cosine functions with different frequencies as well as a subsequent MLP. Finally, the geometric and semantic cues are concatenated and then fed to an MLP, generating the final prompt embeddings $F_p$. Furthermore, we introduce a memory mechanism into CPE to realize a long-term feature dependence. Specifically, we adopt FIFO queue as a memory bank to store the $F_p$ prompt embeddings of previous $T$ frames of local land areas. The memory bank efficiently stores prompt embeddings of the remaining $T-1$ neighboring frames. These embeddings are then aggregated using a weighted sum operation within the CPE. The weights assigned to each stored frame are learnable parameters generated by intermediate layer in CPE, allowing the model to adaptively focus on relevant information. This design effectively reduces noise in the prompted information (\textit{e.g.} prompts may contain prediction errors) while maintaining a lightweight architecture with minimal computational overhead. The memory bank is the core module of the CPE and directly reflects its performance. Finally, $F_p$ is obtained via learning an aggregation of the stored embeddings.

\subsection{Group-wise Lane Prediction (GLP)} 
A lane group refers to a collection of lanes that share common characteristics \textit{e.g.} same style or are directed towards a common goal or destination. Practically, lane groups are essential for path planning and navigation as they help in understanding complex road structures. However, predicting an accurate group-wise lane is challenging since it demands sophisticated semantic analysis and geometrical reasoning. Particularly, localizing the endpoint of a lane instance requires the knowledge of the style and topology changing of the other lanes in the same group. To address this, we propose to use a polygon, namely group polygon to outline the boundary of a lane group. We further introduce an auxiliary task in the network architecture to predict the group polygons. Since all the points of the predicted map elements are located in the group polygons, we propose an additional point-in-polygon loss to facilitate group-wise lane prediction.  

In this section, we introduce the key components regarding to group-wise lane prediction (GLP), including the query design, decoder architecture, and the prediction heads.

\textbf{Queries. }
We design a query combination to flexibly encode structured map information and perform hierarchical bipartite matching for both map element and group polygon learning. Specifically, we extend the hierarchical query scheme in MapTR \cite{liao2022maptr} and customize two set of queries, namely, element queries  $\{q^{l}_i\}^{N_{l}}_{i=0}$ and group polygon queries $\{q^g_i\}^{N_g}_{i=0}$. These two types of queries adopt the same hierarchical query scheme that efficiently encodes instance-level and point-level information. Moreover, we introduce an additional foreground-background (FG) query $q^s$ for the auxiliary task of semantic segmentation.

\textbf{Decoder. } 
All map elements, group polygons, and segmentation masks are simultaneously predicted using a unified Transformer structure. The decoder is composed of several cascaded layers, each incorporating a self-attention module, two cross-attention modules, and an intra-instance self-attention module. The initial self-attention module is designed to enable hierarchical queries to exchange information across the entire feature space. The subsequent cross-attention module facilitates interaction between hierarchical queries and BEV features. To enhance prediction accuracy and spatial consistency, an additional cross-attention module has been innovatively introduced, with contextual prompt embeddings serving as input keys and values to interact with hierarchical queries. Lastly, the intra-instance self-attention module allows for interactions between points within the same instance, thereby improving geometric smoothness. Ultimately, after processing through the decoder, the hierarchical queries are effectively encoded into group-level query embeddings $F_g \in \mathbb{R}^{N_g \times N_p \times 256}$, line-level query embeddings $F_l \in \mathbb{R}^{N_l \times N_p \times 256}$, and a foreground embedding $F_s \in \mathbb{R}^{1 \times N_p \times 256}$.

\textbf{Predictions. } 
For predicting lane lines and lane group polygons, we input both map element and group polygon query embeddings into a shared classification branch and a shared regression branch to facilitate type classification and geometric property regression, respectively. For each predicted instance, the regression branch outputs a vector of dimension $\mathbb{R}^{N_{p} \times 2}$, representing the normalized coordinates of $N_{p}$ points.

Furthermore, to enhance the performance of the classification and regression branches, thereby improving prediction accuracy and accelerating training convergence, we propose a foreground segmentation branch. Instead of directly utilizing the BEV features for segmentation, an individual foreground query $q^{seg} \in \mathbb{R}^{1 \times 256}$ is introduced alongside the hierarchical queries. Following processing by the conventional decoder network and MLP encoding, the foreground query embedding interacts with the BEV feature to generate a segmentation map.

\begin{figure}%
\includegraphics[width=1.0\linewidth,trim={0.0cm 0.2cm 0.0cm 0.0cm},clip]{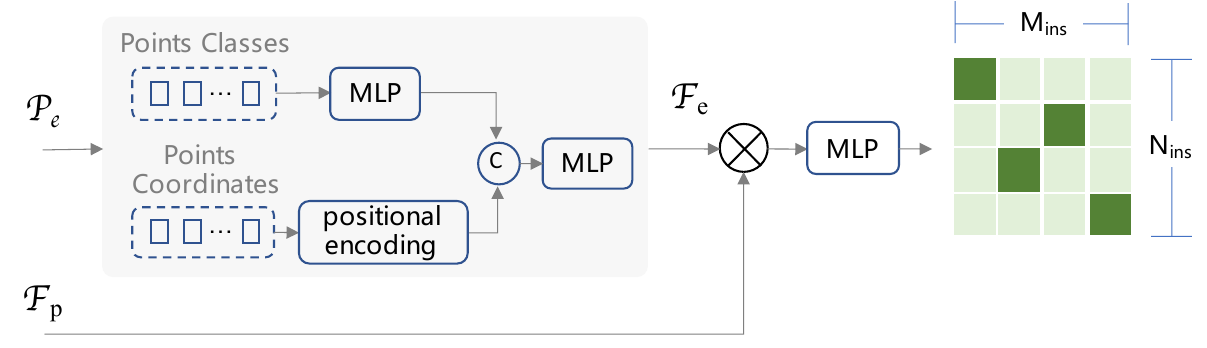}
\caption{Topology prediction. The topology matrix is produced as an additional output of the decoder to indicate the connections between $N_{ins}$ element instances in the current land area and $M_{ins}$ element instances in the contextual land areas.}
\label{fig:topology}
\end{figure}

\subsection{Topology Prediction}
Considering our objective to generate city-scale lane-level map in an end-to-end manner, it is insufficient to solely predict the vectorization of single-frame BEV images. Predicting the topological relationships between frames is indispensable for our task. To address this, we propose to directly predict a topology matrix $\mathcal{M} \in \mathbb{R}^{M_{ins} \times N_{ins}}$ which indicates the connections between $N_{ins}$ element instances in current land area and $M_{ins}$ element instances in the contextual land areas. 

Inspired by \cite{wu20231st}, we formulate topology prediction as a classification task, where the topology matrix is produced as an additional output of the decoder. The architecture is outlined in  Figure~\ref{fig:architecture} and specified in Figure~\ref{fig:topology}. Specifically, we adopt a similar sub-network as the CPE to encoder the predicted map elements, producing an embedding $F_{e} \in \mathbb{R}^{M_{ins}\times N_{p} \times 256}$ that encapsulate both the predicted coordinates and class information of instances. Subsequently, we calculate the correlation of $F_{e}$ and the prompt embeddings $F_{p}$ than adopt a MLP to produce the topology matrix. During inference, lane lines and lane group polygons are aggregated based on their geometric relationships, facilitating the generation of coherent lane group configurations.

\subsection{End-to-End Training}
In the training stage, for each frame, we apply the hierarchical matching scheme in MapTR~\cite{liao2022maptr} to obtain pairs of the map-elements predictions and the ground truths, denoted as $\{\hat{P}^l_i, P^l_i\}$. Meanwhile, we adopt the same matching scheme to obtain pairs of the predicted group polygons and the ground truths, denoted as $\{\hat{P}^g_j, P^g_j\}$.

Based on the matching results and the correspondence between map elements and their group polygons, we employ multiple task-specific losses to train our proposed DuMapNet in an end-to-end manner:
\begin{equation}
\label{equ2}
\mathcal{L} = \alpha \mathcal{L}_{l} + \beta \mathcal{L}_{t} + \lambda \mathcal{L}_{g} + \eta \mathcal{L}_{gl} + \mu \mathcal{L}_{s}
\end{equation}
where $\mathcal{L}_l$, $\mathcal{L}_t$ denotes the loss for learning map element and topology, respectively. Besides, $\mathcal{L}_{g}$ denotes the loss for learning group polygons while $\mathcal{L}_{gl}$ is an additional point-in-polygon loss to facilitate group-wise elements organization. Finally, we introduce a foreground segmentation loss $\mathcal{L}_{s}$ to enhance lane point localization. $\alpha, \beta, \lambda, \eta, \mu$ are hyperparameters that strike a balance between difference losses. Next, we provide detailed illustrations for each type of loss function. 

\textbf{Map Elements Learning. }
For each map element instance, we employ aligned Focal Loss $\mathcal{L}_{cls}$ for style classification and an L1 regression loss $\mathcal{L}_{L1}$ for point localization, respectively. Here we follow MapTR \cite{liao2022maptr} and further apply a direction loss $\mathcal{L}_{dir}$ to align the direction of the predicted lane segments with the ground truth. The loss for training the map elements is thus denoted as:
\begin{equation}
\label{equ3}
\mathcal{L}_{l} = \mathcal{L}_{cls} + \mathcal{L}_{reg} + \mathcal{L}_{dir}
\end{equation}

Furthermore, inspired by Stable DINO \cite{liu2023detection}, we adapt the aligned Focal Loss to enhance the alignment between classification score and localization quality in $\mathcal{L}_{cls}$. As shown in Equation.~(\ref{Lcls}), L1 distance of the $i$-th matched pairs between prediction $\hat{P}^l_i$ and corresponding ground truth $P^l_i$ is used as a positional metric to supervise the training probabilities of positive examples. The classification loss is thus formulated as:

\begin{equation}
\label{Lcls}
\mathcal{L}_{cls} = \sum_{i=1}^{N_{pos}} (|d_i - s_i|^\gamma) BCE(s_i, d_i) + \sum_{i=1}^{N_{neg}} s_i^\gamma BCE(s_i, 0)
\end{equation}

\begin{equation}
\label{equ5}
d_i = || \hat{P}^l_i - P^l_i ||_{1}
\end{equation}
where $s_i$ is the probability for the $i$-th predicted map element. $N_{pos}$ and $N_{neg}$ denote the number of positive and negative elements, respectively. Moreover, as for $\mathcal{L}_{t}$, we apply the same loss function as $\mathcal{L}_{l}$ for learning group polygons.

\textbf{Topology Learning. }
We define the topological relationship as a 2-class classification task \textit{i.e.} connected or not. As the number of connected ones is significantly less than the number of unconnected ones, we apply a focal loss to supervise the topological association matrix prediction.

\textbf{Group-guided Auxiliary Supervision. }
We leverage the group polygons to provide auxiliary supervision for learning high-quality group-wise lane lines. Specifically, this auxiliary supervision is designed based on the following observations: all the points of the lane lines should be located inside or on the boundary of their corresponding group polygons. We thus propose a point-in-polygon loss as:

\begin{equation}
\label{equ6}
\mathcal{L}_{gl}(\hat{P}^l_i, P^g_j)= \sum_{\substack{p \in \hat{P}^l_i\ and \\
outside\ P^g_j}} D(p, P^g_j)
\end{equation}
where $D(p, P^g_j)$ is the closest distance from point $p$ to any edge of the lane group polygon $P^g_j$. During the training stage, we adopt the ground-truth group polygons to punish predicted points located outside their group polygon. During the inference stage, we simply employ the predicted group polygons to check map elements' locations.

\textbf{Segmentation-guided Auxiliary Supervision. }
To improve the accuracy of vectorized predictions, we choose a combination of Binary Cross Entropy and Dice loss to calculate the loss between the predicted foreground mask $F_s \in R^{H \times W}$ and the ground truth mask $F_m \in R^{H \times W}$ generated from ground truth lane lines:
\begin{equation}
\label{equ7}
\mathcal{L}_{s} = BCE(F_s, F_m) + \mathcal{L}_{dice}(F_s, F_m)
\end{equation}

\section{Experiments}
\subsection{Experimental Settings}

\textbf{Datasets. }
As aforementioned, DuMapNet has been deployed at Baidu Maps, supporting over $360$ cities. To evaluate the effectiveness of DuMapNet, we have collected a large-scale real-world dataset, DuLD, consisting of bird's-eye view (BEV) images and ground truth data from six cities: Beijing, Guangzhou, Changchun, Changzhou, Chongqing, and Leshan. These cities were chosen for their varied urban scales and geographic features. The dataset from Beijing, Guangzhou, Changchun, and Changzhou was divided into a training set and a validation set in a $9:1$ ratio. Meanwhile, data from Chongqing and Leshan were used as the test set to evaluate the model's performance. Statistically, DuLD contains $134,524$ images, spanning $8,072$ kilometers, with each image at a resolution of $1536 \times 1536$ pixels. More details can be found in Table~\ref{table:dataset}. Importantly, to investigate the benefits of larger-scale data, we introduce DuLD-L, a dataset with one million paired images and corresponding ground truths, and evaluate DuMapNet's performance on this expanded dataset.

\textbf{Evaluation Metrics. }
We adopt recall (R) and precision (P) to assess the quality of map construction at the instance level. The evaluation considers category consistency, endpoint distance, and overlap to determine if a pair of lane instances from ground truth and prediction match. Moreover, category consistency and IoU ($IoU > 0.5$) are used for closed-shape elements such as crosswalks. Specifically, category consistency mandates that the instances belong to the same category. Endpoint distance requires the L2 distance between the start points and end points of the instances should be less than 3 meters, respectively. Overlap considers the parallel distance between instances. When calculating overlap, both prediction instances and ground truth are first divided into multiple segments with 1-meter intervals. Then, the projection distance between segments is calculated. If the proportion of segments with projection distance smaller than the threshold $d=\{0.5, 1\}m$ exceeds the threshold $r=\{0.5, 0.8\}$, the prediction instance will be considered as true positive (TP). In the following experiments, we use $R@P_{d,r}=p$ to represent the recall at $p$ precision with threshold $d$ and $r$. Lower values of threshold $d$ and higher values of threshold $r$ signify more stringent precision requirements.

\textbf{Implementation Details. }
Our model is trained using 16 NVIDIA Tesla V100 GPUs, with a batch size of 16. The AdamW~\cite{loshchilov2017adamw} optimizer is employed with a weight decay of 0.01 and the initial learning rate is set to $6\times10^{-4}$ with cosine decay. The input images have a resolution of $768\times768$ pixels. For our architecture, we employ ResNet50~\cite{he2016resnet} and HRNet48~\cite{wang2020hrnet} as backbones. The default number of instance queries, point queries and decoder layers are $50$, $50$ and $6$, respectively. As for hyper-parameters of loss weight, we set $\alpha$, $\beta$, $\lambda$, $\eta$ and $\mu$ to $1$, $1$, $0.2$, $0.15$ and $100$, respectively. The inference time is measured on a single NVIDIA Tesla V100 GPU with batch size $1$.

\subsection{Evaluation}
\textbf{Comparison with Baselines. }
We compare our DuMapNet with segmentation-based method~\cite{tao2020hmsa} and other vectorization-based methods~\cite{liao2022maptr,zhang2023gemap}. As shown in Table~\ref{tab:sota1}, vectorization-based methods achieve better results without complex post-processing logic. In particular, our DuMapNet outperforms the existing state-of-the-art method by a large margin ($+2.66\%$ when $P_{1,0.8}=90\%$) under the same setting of ResNet50, indicating the effectiveness of our method. Surprisingly, our method achieves a further improvement of $4.99\%$ by replacing the backbone with HRNet48 to obtain enhanced feature representation. Furthermore, our method achieves $73.28\%$ recall on one million of training data, demonstrating that as the training data volume increases, our method can achieve greater advantages.

The quantitative comparisons of different evaluation thresholds are summarized in Table~\ref{tab:sota2}. From the results, we observe that DuMapNet consistently brings significant improvements. Taking $R@P_{0.5,0.8}=90\%$ as an example, DuMapNet achieves better performance with $+3.00\% \sim 5.04\%$ recall gains on DuLD, which underscores our method's ability to achieve superior geometric precision and maintain category consistency. Particularly, as the projection distance decreases, the performance of all approaches experiences a significant decline, yet our DuMapNet shows a slighter drop, indicating that our method is more robust and maintains superior performance at higher precision levels.

\begin{table}
\centering
\caption{Statistics of DuLD dataset.}
\begin{center}
\scalebox{1.0} {
\begin{tabular}{c|cc|ll}
\toprule
Dataset & Mileage (km) & Image & \multicolumn{2}{c}{City} \\
\midrule
Train & \multirow{2}*{7,922} & 118,822 & Beijing & Guangzhou \\
Val &  & 13,202 & Changchun & Changzhou \\
\midrule
Test & 150 & 2,500 & Chongqing & Leshan \\
\midrule
All & 8,072 & 134,524 & \multicolumn{2}{c}{-} \\
\bottomrule
\end{tabular}
}
\end{center}
\label{table:dataset}
\end{table}

\textbf{Ablation Studies.} In this section, extensive ablation studies are conducted to systematically evaluate the key designs of our DuMapNet. As shown in Table~\ref{table:ablation_1}, Group I is the baseline without a series of designs. From Group I and II, it is proved that adding the task-aligned supervision can bring slight improvements by fostering better synergy between geometric learning and category identification. Further analysis between Groups II and III reveals that incorporating intra-instance self-attention results in a $0.99$ improvement under $P_{1,0.8}=95\%$. The transition from Group III to Group IV examines the impact of adding segmentation-guided auxiliary supervision, showing marked improvements in all metrics, especially at higher precision levels. This outcome is expected as the segmentation branch contributes to more fine-grained pixel-level modeling, enhancing semantic understanding and refining prediction accuracy. The results from Group V highlight the significant role of the proposed contextual prompts encoder (CPE) module, showing a notable increase of $2.74\%$ in recall at a high accuracy level ($P_{1,0.8}=95\%$). These findings demonstrate that CPE, by leveraging spatial prediction results from the area surrounding the current BEV image, significantly enhances the geometric and category consistency of predictions across a broader receptive field.

\begin{table*}
\caption{Comparisons with state-of-the-art methods on DuLD test set at different precision levels. R50 and HR48 correspond to ResNet50~\cite{he2016resnet} and HRNet48~\cite{wang2020hrnet}, respectively. FPSs are measured on the same machine with NVIDIA
Tesla V100.}
\centering
\begin{tabular}{c|cc|ccc|c}
\toprule
\textbf{Method} & \textbf{Backbone} & \textbf{Training Set} & \textbf{$R@P_{1,0.8}=80\%$} & \textbf{$R@P_{1,0.8}=90\%$} & \textbf{$R@P_{1,0.8}=95\%$} & FPS\\
\midrule
HMSA \cite{tao2020hmsa} + post-processing & HR48 & DuLD & 58.48 & - & - & - \\
\midrule
MapTR \cite{liao2022maptr} & R50 & DuLD & 69.56 & 58.58 & 39.49 & 29.7 \\
GeMap \cite{zhang2023gemap} & R50 & DuLD & 71.96 & 60.45 & 39.32 & 29.3 \\
\midrule
DuMapNet & R50 & DuLD & 74.61 & 63.11 & 43.27 & 27.9 \\
DuMapNet & HR48 & DuLD & 77.34 & 68.10 & 54.21 & 26.6 \\
DuMapNet & HR48 & DuLD-L & \textbf{83.40} & \textbf{73.28} & \textbf{61.24} & 26.6 \\
\bottomrule
\end{tabular}
\label{tab:sota1}
\end{table*}

\begin{table*}
\caption{Comparisons with state-of-the-art methods on DuLD test set under different thresholds $d$ and $r$, where $d$ represents the projection distance threshold and $r$ represents the proportion of segments threshold.}
\centering
\begin{tabular}{c|cc|cccc}
\toprule
\textbf{Method} & \textbf{Backbone} & \textbf{Training Set} & \textbf{$R@P_{1,0.8}=90\%$} & \textbf{$R@P_{1,0.5}=90\%$} & \textbf{$R@P_{0.5,0.8}=90\%$} & \textbf{$R@P_{0.5,0.5}=90\%$} \\
\midrule
MapTR \cite{liao2022maptr} & R50 & DuLD & 58.58 & 59.29 & 54.29 & 56.5 \\
GeMap \cite{zhang2023gemap} & R50 & DuLD & 60.45 & 61.0 & 56.33 & 58.15 \\
\midrule
DuMapNet & R50 & DuLD & 63.11 & 63.56 & 59.33 & 60.87 \\
DuMapNet & HR48 & DuLD & 68.10 & 68.91 & 65.47 & 66.91 \\
DuMapNet & HR48 & DuLD-L & \textbf{73.28} & \textbf{73.98} & \textbf{71.66} & \textbf{72.41} \\
\bottomrule
\end{tabular}
\label{tab:sota2}
\end{table*}

\begin{table*}
\small
\centering
\caption{Effects of core components of DuMapNet.}
\scalebox{0.95} {
\begin{tabular}{c|cccc|ccc}
\toprule
\multirow{2}*{\textbf{Group}} & \textbf{Task-aligned} & \textbf{Intra-instance}  & \textbf{Segmentation-guided} & \multirow{2}*{\textbf{CPE}}  & \multirow{2}*{\textbf{$R@P_{1,0.8}=80\%$}} & \multirow{2}*{\textbf{$R@P_{1,0.8}=90\%$}} & \multirow{2}*{\textbf{$R@P_{1,0.8}=95\%$}} \\
&\textbf{Supervision}&\textbf{Self-attention}&\textbf{Auxiliary Supervision}&&&& \\
\midrule
I & & & & & 73.93 & 63.11 & 49.42 \\
II & \checkmark & & & & 74.04 & 63.61 & 49.72 \\
III & \checkmark & \checkmark & & & 76.95 & 64.60 & 49.87 \\
IV & \checkmark & \checkmark & \checkmark & & 77.18 & 65.81 & 51.47 \\
V & \checkmark & \checkmark & \checkmark & \checkmark & 77.34 & 68.10 & 54.21 \\
\bottomrule
\end{tabular}
}
\label{table:ablation_1}
\end{table*}

\begin{table*}
\caption{Comparisons with state-of-the-art methods on the additional test sets are conducted to assess the generalization performance. The additional test sets consist of data from five cities, including Harbin, Hangzhou, Xi'an, Zhongshan, and Shanghai. The metric is $R@P_{1,0.8}=80\%$.}
\centering
\begin{tabular}{c|cc|ccccc}
\toprule
\textbf{Method} & \textbf{Backbone} & \textbf{Training Set} & \textbf{Harbin} & \textbf{Hangzhou} & \textbf{Xi'an} & \textbf{Zhongshan} & \textbf{Shanghai} \\
\midrule
MapTR\cite{liao2022maptr} & R50 & DuLD & 73.30 & 66.30 & 71.76 & 73.55 & 69.66  \\
GeMap\cite{zhang2023gemap} & R50 & DuLD & 75.80 & 69.07 & 72.03 & 75.19 & 71.81 \\
\midrule
DuMapNet & R50 & DuLD & 76.12 & 73.35 & 76.11 & 77.30 & 73.31 \\
DuMapNet & HR48 & DuLD & 78.73 & 76.19 & 77.50 & 77.33 & 76.66 \\
DuMapNet & HR48 & DuLD-L & \textbf{83.99} & \textbf{82.10} & \textbf{82.06} & \textbf{82.55} & \textbf{84.38} \\
\bottomrule
\end{tabular}
\label{tab:generalization}
\end{table*}

\textbf{Analysis of generalization. }
To further demonstrate the generalization of our method, five cities are additionally selected as the test set. These cities are distributed across various regions, such as Harbin in northeastern China and Xi'an in northwestern China, and exhibit diverse sizes, with Shanghai being a large first-tier city, while Zhongshan is a second-tier city. Finally, a total of $5,000$ images were collected for evaluation. The experimental results are presented in Table~\ref{tab:generalization}. On the one hand, DuMapNet outperforms the existing state-of-the-art methods on all city test sets, demonstrating the effectiveness of our approach. On the other hand, DuMapNet shows superior generalization with less fluctuation in performance across the five cities. For example, the maximum deviation of DuMapNet-R50 across the five cities is $3.99\%$, while the second-best model, \textit{i.e.}, GeMap has a maximum deviation of $6.73\%$.

\begin{figure*}
\includegraphics[width=1.0\linewidth]{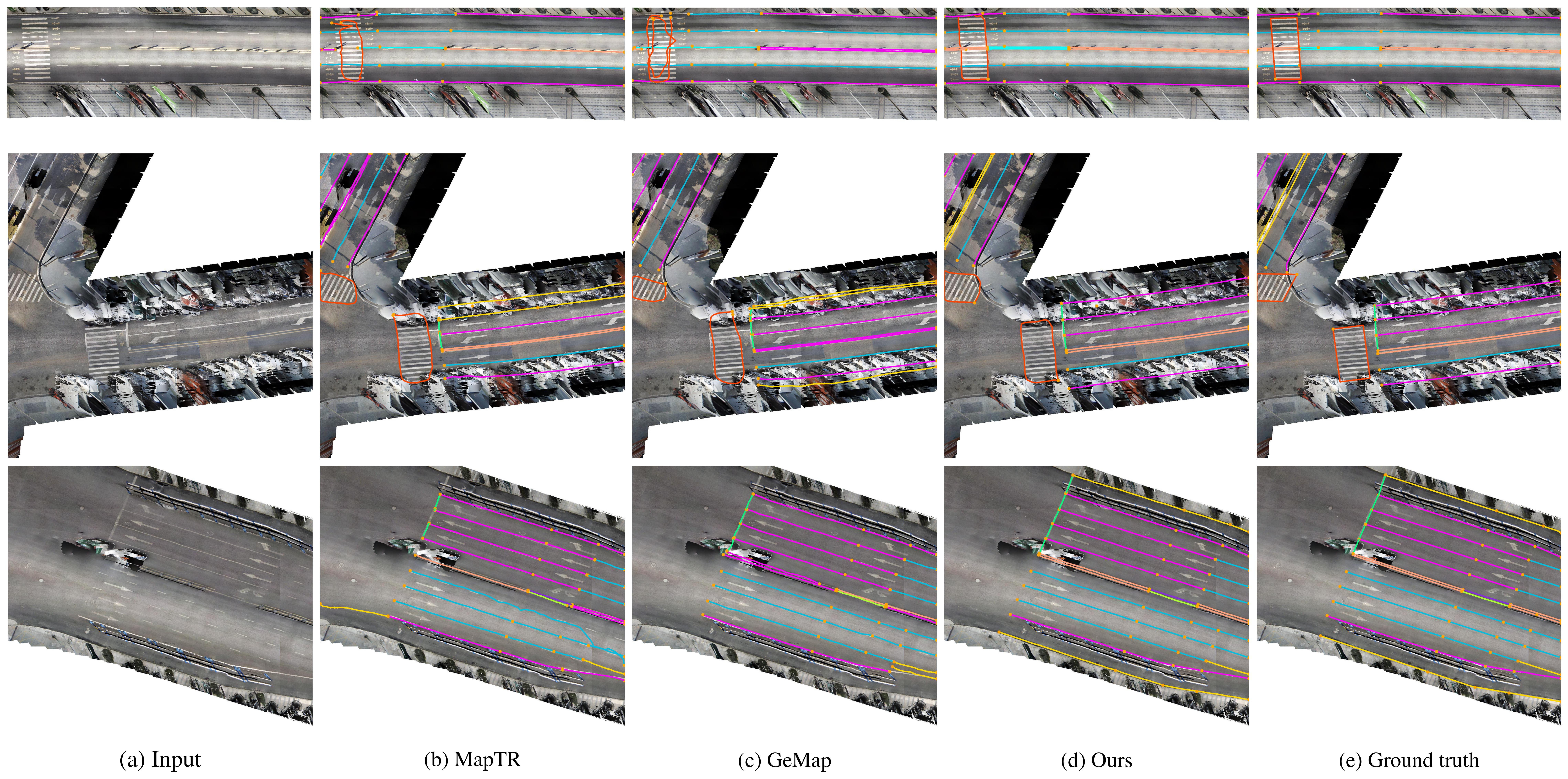}
\caption{Comparisons of our method with state-of-the-art models in lane-level map generation.}
\label{fig:comparisons}
\end{figure*}

\begin{figure*}
\includegraphics[width=1.0\linewidth]{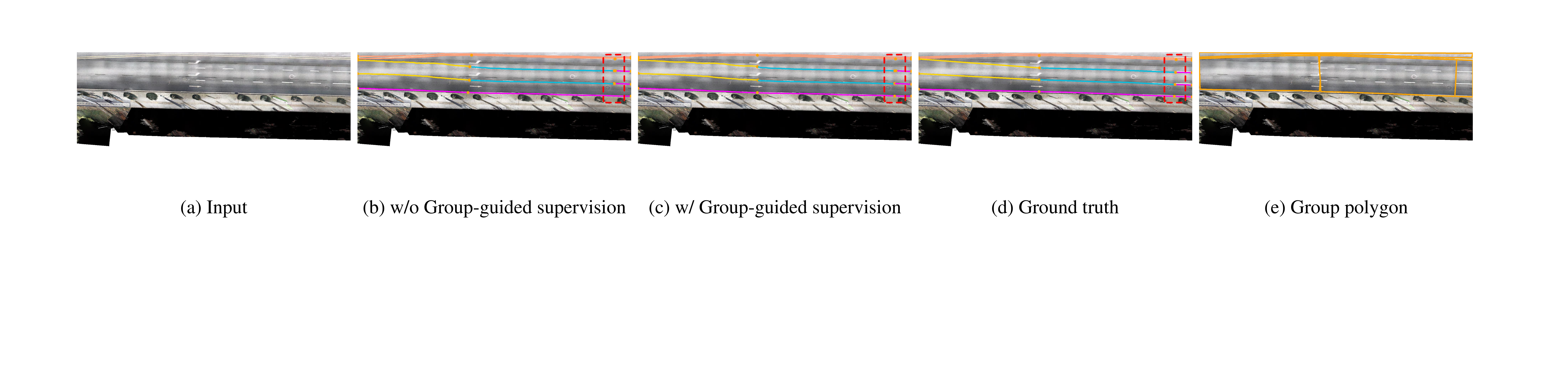}
\caption{Qualitative visualization of the proposed group-guided supervision.}
\label{fig:ab}
\end{figure*}

\begin{figure*}
\includegraphics[width=1.0\linewidth]{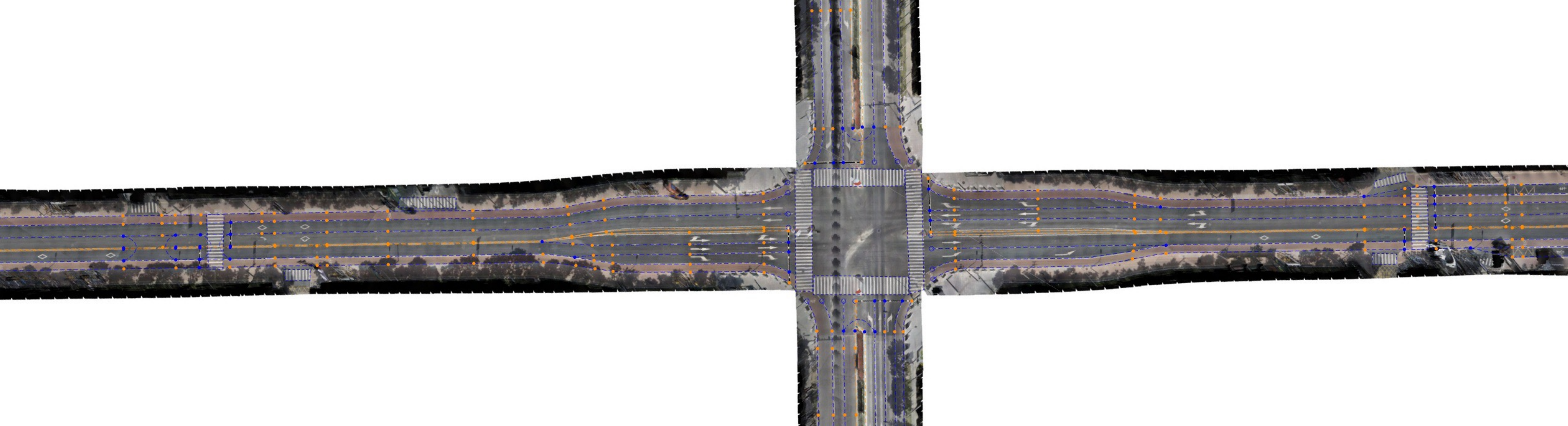}
\caption{Qualitative visualization of urban scene.}
\label{fig:global}
\end{figure*}

\subsection{Visualization}
Qualitative results from the DuLD dataset are presented in Figure~\ref{fig:comparisons} and Figure~\ref{fig:ab}. DuMapNet not only performs well in simple scenes but also predicts high-quality vectorized map elements in complex scenarios like intersections, road wear and occlusions. As shown in Figure~\ref{fig:comparisons}, DuMapNet exhibits significant advantages in terms of lane recall, lane accuracy, and endpoint accuracy. For example, as demonstrated in the second row of Figure~\ref{fig:comparisons}, DuMapNet precisely captures both the geometry and category of lane lines in occlusion scenes, avoiding unnecessary lane line predictions. In addition, Figure~\ref{fig:ab} provides a visual comparison that underscores the effectiveness of group-guided supervision in enabling accurate prediction of endpoint positions, even in scenarios with subtle visual differences, such as the dashed line and the segment of solid line. Group-guided supervision also ensures that the endpoints of lane lines within a lane group are correctly aligned. Furthermore, as illustrated in Figure~\ref{fig:global}, the implementation of the topology prediction module enables the end-to-end generation of comprehensive urban scenes at a global level. 

\section{Related Work}
Here we briefly review the closely related work in the fields of map construction and lane detection.

\subsection{Map Construction}
With the development of deep learning and BEV perception \cite{huang2021bevdet}, map construction is transitioning from a labor-intensive annotation task to a model-based dense prediction challenge. Segmentation-based methods \cite{pan2020vpn, li2022bevformer, hu2021fiery, peng2023bevsegformer, harley2022simple} generate rasterized map by performing BEV semantic segmentation. To build vectorized maps, HDMapNet \cite{li2022hdmapnet} adopts a two-stage approach of segmentation followed by post-processing to generate vectorized instances. As the first end-to-end framework, VectorMapNet \cite{liu2023vectormapnet} utilizes an auto-regressive decoder to predict points sequentially. MapTR \cite{liao2022maptr} proposes a unified shape modeling method based on a parallel end-to-end framework, which has been followed by many works \cite{qiao2023bemapnet, ding2023pivotnet, qiao2023machmap}. MapVR \cite{zhang2023mpvr} applies differentiable rasterization to vectorized outputs to performs precise and geometry-aware supervision. MapTRv2 \cite{liao2023maptrv2} further introduces auxiliary one-to-many matching and auxiliary dense supervision to speedup convergence. BeMapNet \cite{qiao2023bemapnet} adopts a parameterized paradigm and constructs map elements as piecewise Bezier curves. PivotNet \cite{ding2023pivotnet} utilizes a dynamic number of pivotal points to model map elements, preventing the loss of essential details. Different from the existing works, our proposed DuMapNet leverages neighboring map elements as prompts to guide the generation of map elements in the current frame, which can enhance the spatial consistency of the map elements.

\subsection{Lane Detection}
Lane detection plays a critical role in detecting lane elements in road scenes and can be considered as a subtask of map construction. LaneATT \cite{tabelini2021laneatt} utilizes an anchor-based deep lane detection model. CondLaneNet \cite{liu2021condlanenet} adopts a conditional lane detection model based on conditional convolution and row-wise formulation. GANet \cite{wang2022ganet} formulates lane detection as a keypoint estimation and association problem. BezierLaneNet \cite{feng2022bezierlaneNet} proposes a parametric Bezier curve-based method, which can model the geometric shapes of lane lines. PersFormer \cite{chen2022persformer} utilizes a transformer-based spatial feature transformation module and unify 2D and 3D lane detection simultaneously. Different from these methods that primarily focus on lane elements, our proposed DuMapNet models map elements in a unified vectorized form, which can detect open-shape map elements such as lanes, as well as closed-shape elements like crosswalks.

\section{Discussion}
Before the deployment of DuMapNet, Baidu Maps relied heavily on labor-intensive manual annotation processes that involved segmentation techniques and complex post-processing logic. This approach significantly increased operational costs and decreased efficiency. With the introduction of DuMapNet, now operational in over $360$ cities, production efficiency has seen a twenty-fold improvement, leading to a remarkable $95\%$ reduction in costs.

Despite DuMapNet's impressive achievements, several challenging issues remain unresolved and require further investigation. For instance, the model struggles in scenarios with extensive static obstructions, such as long stretches of road with parked vehicles. Such conditions disrupt performance because the lack of visible road surface markings compromises the effectiveness of the contextual prompts encoder. To address this challenge, integrating multi-source data may be an effective approach. In addition, generating qualified map data from low-precision sources, such as crowdsourced data, presents an intriguing challenge that merits deeper exploration in future work. Currently, leveraging the high timeliness, broad coverage, and low cost of crowdsourced data for map updates represents a more reasonable paradigm. For example, crowdsourced data can provide timely updates for elements with lower accuracy requirements, such as style changes, or for dynamic changes like construction or temporary road closures. 

\section{Conclusions}
In this paper, we present an effective industrial solution for city-scale lane-level map generation. Specifically, we reformulate this task as a vectorization modeling task that takes bird's-eye-view (BEV) images as input and outputs standardized, vectorized map elements and their topology in an end-to-end paradigm. We pioneer organize the lane group using a learning-based methodology and address it through the proposed group-wise lane prediction (GLP) system that outputs vectorized results of lane groups by applying mutual constraints between lane group polygons and lane lines, thereby eliminating the need for intricate post-processing logic. To improve the generalization in challenging scenarios, such as road wear and occlusions, as well as to improve the continuity of vectorization results across frames, we present the contextual prompts encoder (CPE) module, which leverages the spatial prediction results from the surrounding area of the current BEV image as contextual information. Extensive experiments conducted on the collected large-scale real-world dataset from Baidu Maps demonstrate the superiority of DuMapNet. The successful deployment of DuMapNet at Baidu Maps has significantly improved its performance. Since its launch in June 2023, DuMapNet served over $360$ cities while bringing a $95\%$ reduction in costs.

\section{Acknowledgments}
This work was supported in part by the National Natural Science Foundation of China (52102464), Beijing Natural Science Foundation (L231008), and Young Elite Scientist Sponsorship Program By BAST (BYESS2022153). 

\clearpage
\balance
\bibliographystyle{ACM-Reference-Format}
\bibliography{ref}


\begin{thebibliography}{40}


\ifx \showCODEN    \undefined \def \showCODEN     #1{\unskip}     \fi
\ifx \showDOI      \undefined \def \showDOI       #1{#1}\fi
\ifx \showISBNx    \undefined \def \showISBNx     #1{\unskip}     \fi
\ifx \showISBNxiii \undefined \def \showISBNxiii  #1{\unskip}     \fi
\ifx \showISSN     \undefined \def \showISSN      #1{\unskip}     \fi
\ifx \showLCCN     \undefined \def \showLCCN      #1{\unskip}     \fi
\ifx \shownote     \undefined \def \shownote      #1{#1}          \fi
\ifx \showarticletitle \undefined \def \showarticletitle #1{#1}   \fi
\ifx \showURL      \undefined \def \showURL       {\relax}        \fi
\providecommand\bibfield[2]{#2}
\providecommand\bibinfo[2]{#2}
\providecommand\natexlab[1]{#1}
\providecommand\showeprint[2][]{arXiv:#2}

\bibitem[Ammar~Abbas and Zisserman(2019)]%
        {ammar2019geometric}
\bibfield{author}{\bibinfo{person}{Syed Ammar~Abbas} {and} \bibinfo{person}{Andrew Zisserman}.} \bibinfo{year}{2019}\natexlab{}.
\newblock \showarticletitle{A geometric approach to obtain a bird's eye view from an image}. In \bibinfo{booktitle}{\emph{Proceedings of the IEEE/CVF international conference on computer vision workshops}}. \bibinfo{pages}{0--0}.
\newblock


\bibitem[Chen et~al\mbox{.}(2022)]%
        {chen2022persformer}
\bibfield{author}{\bibinfo{person}{Li Chen}, \bibinfo{person}{Chonghao Sima}, \bibinfo{person}{Yang Li}, \bibinfo{person}{Zehan Zheng}, \bibinfo{person}{Jiajie Xu}, \bibinfo{person}{Xiangwei Geng}, \bibinfo{person}{Hongyang Li}, \bibinfo{person}{Conghui He}, \bibinfo{person}{Jianping Shi}, \bibinfo{person}{Yu Qiao}, {et~al\mbox{.}}} \bibinfo{year}{2022}\natexlab{}.
\newblock \showarticletitle{Persformer: 3d lane detection via perspective transformer and the openlane benchmark}. In \bibinfo{booktitle}{\emph{European Conference on Computer Vision}}. Springer, \bibinfo{pages}{550--567}.
\newblock


\bibitem[Ding et~al\mbox{.}(2023)]%
        {ding2023pivotnet}
\bibfield{author}{\bibinfo{person}{Wenjie Ding}, \bibinfo{person}{Limeng Qiao}, \bibinfo{person}{Xi Qiu}, {and} \bibinfo{person}{Chi Zhang}.} \bibinfo{year}{2023}\natexlab{}.
\newblock \showarticletitle{Pivotnet: Vectorized pivot learning for end-to-end hd map construction}. In \bibinfo{booktitle}{\emph{Proceedings of the IEEE/CVF International Conference on Computer Vision}}. \bibinfo{pages}{3672--3682}.
\newblock


\bibitem[Fang et~al\mbox{.}(2021)]%
        {fang2021ssml}
\bibfield{author}{\bibinfo{person}{Xiaomin Fang}, \bibinfo{person}{Jizhou Huang}, \bibinfo{person}{Fan Wang}, \bibinfo{person}{Lihang Liu}, \bibinfo{person}{Yibo Sun}, {and} \bibinfo{person}{Haifeng Wang}.} \bibinfo{year}{2021}\natexlab{}.
\newblock \showarticletitle{SSML: Self-Supervised Meta-Learner for En Route Travel Time Estimation at Baidu Maps}. In \bibinfo{booktitle}{\emph{Proceedings of the 27th ACM SIGKDD Conference on Knowledge Discovery \& Data Mining}}. \bibinfo{pages}{2840--2848}.
\newblock


\bibitem[Fang et~al\mbox{.}(2020)]%
        {fang2020constgat}
\bibfield{author}{\bibinfo{person}{Xiaomin Fang}, \bibinfo{person}{Jizhou Huang}, \bibinfo{person}{Fan Wang}, \bibinfo{person}{Lingke Zeng}, \bibinfo{person}{Haijin Liang}, {and} \bibinfo{person}{Haifeng Wang}.} \bibinfo{year}{2020}\natexlab{}.
\newblock \showarticletitle{ConSTGAT: Contextual Spatial-Temporal Graph Attention Network for Travel Time Estimation at Baidu Maps}. In \bibinfo{booktitle}{\emph{Proceedings of the 26th ACM SIGKDD International Conference on Knowledge Discovery \& Data Mining}}. \bibinfo{pages}{2697--2705}.
\newblock


\bibitem[Feng et~al\mbox{.}(2022)]%
        {feng2022bezierlaneNet}
\bibfield{author}{\bibinfo{person}{Zhengyang Feng}, \bibinfo{person}{Shaohua Guo}, \bibinfo{person}{Xin Tan}, \bibinfo{person}{Ke Xu}, \bibinfo{person}{Min Wang}, {and} \bibinfo{person}{Lizhuang Ma}.} \bibinfo{year}{2022}\natexlab{}.
\newblock \showarticletitle{Rethinking efficient lane detection via curve modeling}. In \bibinfo{booktitle}{\emph{Proceedings of the IEEE/CVF Conference on Computer Vision and Pattern Recognition}}. \bibinfo{pages}{17062--17070}.
\newblock


\bibitem[Harley et~al\mbox{.}(2022)]%
        {harley2022simple}
\bibfield{author}{\bibinfo{person}{Adam~W Harley}, \bibinfo{person}{Zhaoyuan Fang}, \bibinfo{person}{Jie Li}, \bibinfo{person}{Rares Ambrus}, {and} \bibinfo{person}{Katerina Fragkiadaki}.} \bibinfo{year}{2022}\natexlab{}.
\newblock \showarticletitle{A simple baseline for bev perception without lidar}.
\newblock \bibinfo{journal}{\emph{arXiv e-prints}} (\bibinfo{year}{2022}), \bibinfo{pages}{arXiv--2206}.
\newblock


\bibitem[He et~al\mbox{.}(2016)]%
        {he2016resnet}
\bibfield{author}{\bibinfo{person}{Kaiming He}, \bibinfo{person}{Xiangyu Zhang}, \bibinfo{person}{Shaoqing Ren}, {and} \bibinfo{person}{Jian Sun}.} \bibinfo{year}{2016}\natexlab{}.
\newblock \showarticletitle{Deep residual learning for image recognition}. In \bibinfo{booktitle}{\emph{Proceedings of the IEEE conference on computer vision and pattern recognition}}. \bibinfo{pages}{770--778}.
\newblock


\bibitem[Hu et~al\mbox{.}(2021)]%
        {hu2021fiery}
\bibfield{author}{\bibinfo{person}{Anthony Hu}, \bibinfo{person}{Zak Murez}, \bibinfo{person}{Nikhil Mohan}, \bibinfo{person}{Sof{\'\i}a Dudas}, \bibinfo{person}{Jeffrey Hawke}, \bibinfo{person}{Vijay Badrinarayanan}, \bibinfo{person}{Roberto Cipolla}, {and} \bibinfo{person}{Alex Kendall}.} \bibinfo{year}{2021}\natexlab{}.
\newblock \showarticletitle{Fiery: Future instance prediction in bird's-eye view from surround monocular cameras}. In \bibinfo{booktitle}{\emph{Proceedings of the IEEE/CVF International Conference on Computer Vision}}. \bibinfo{pages}{15273--15282}.
\newblock


\bibitem[Huang et~al\mbox{.}(2021)]%
        {huang2021bevdet}
\bibfield{author}{\bibinfo{person}{Junjie Huang}, \bibinfo{person}{Guan Huang}, \bibinfo{person}{Zheng Zhu}, \bibinfo{person}{Yun Ye}, {and} \bibinfo{person}{Dalong Du}.} \bibinfo{year}{2021}\natexlab{}.
\newblock \showarticletitle{Bevdet: High-performance multi-camera 3d object detection in bird-eye-view}.
\newblock \bibinfo{journal}{\emph{arXiv preprint arXiv:2112.11790}} (\bibinfo{year}{2021}).
\newblock


\bibitem[Huang et~al\mbox{.}(2022)]%
        {dueta2022}
\bibfield{author}{\bibinfo{person}{Jizhou Huang}, \bibinfo{person}{Zhengjie Huang}, \bibinfo{person}{Xiaomin Fang}, \bibinfo{person}{Shikun Feng}, \bibinfo{person}{Chen Xuyi}, \bibinfo{person}{Liu Jiaxiang}, \bibinfo{person}{Yuan Haitao}, {and} \bibinfo{person}{Haifeng Wang}.} \bibinfo{year}{2022}\natexlab{}.
\newblock \showarticletitle{DuETA: Traffic Congestion Propagation Pattern Modeling via Efficient Graph Learning for ETA Prediction at Baidu Maps}. In \bibinfo{booktitle}{\emph{Proceedings of the 31st ACM International Conference on Information and Knowledge Management}}.
\newblock


\bibitem[Jia et~al\mbox{.}(2022)]%
        {jia2022visual}
\bibfield{author}{\bibinfo{person}{Menglin Jia}, \bibinfo{person}{Luming Tang}, \bibinfo{person}{Bor-Chun Chen}, \bibinfo{person}{Claire Cardie}, \bibinfo{person}{Serge Belongie}, \bibinfo{person}{Bharath Hariharan}, {and} \bibinfo{person}{Ser-Nam Lim}.} \bibinfo{year}{2022}\natexlab{}.
\newblock \showarticletitle{Visual prompt tuning}. In \bibinfo{booktitle}{\emph{European Conference on Computer Vision}}. Springer, \bibinfo{pages}{709--727}.
\newblock


\bibitem[Kim et~al\mbox{.}(2021)]%
        {kim2021scan}
\bibfield{author}{\bibinfo{person}{Giseop Kim}, \bibinfo{person}{Sunwook Choi}, {and} \bibinfo{person}{Ayoung Kim}.} \bibinfo{year}{2021}\natexlab{}.
\newblock \showarticletitle{Scan context++: Structural place recognition robust to rotation and lateral variations in urban environments}.
\newblock \bibinfo{journal}{\emph{IEEE Transactions on Robotics}} \bibinfo{volume}{38}, \bibinfo{number}{3} (\bibinfo{year}{2021}), \bibinfo{pages}{1856--1874}.
\newblock


\bibitem[Kirillov et~al\mbox{.}(2023)]%
        {kirillov2023segment}
\bibfield{author}{\bibinfo{person}{Alexander Kirillov}, \bibinfo{person}{Eric Mintun}, \bibinfo{person}{Nikhila Ravi}, \bibinfo{person}{Hanzi Mao}, \bibinfo{person}{Chloe Rolland}, \bibinfo{person}{Laura Gustafson}, \bibinfo{person}{Tete Xiao}, \bibinfo{person}{Spencer Whitehead}, \bibinfo{person}{Alexander~C Berg}, \bibinfo{person}{Wan-Yen Lo}, {et~al\mbox{.}}} \bibinfo{year}{2023}\natexlab{}.
\newblock \showarticletitle{Segment anything}.
\newblock \bibinfo{journal}{\emph{arXiv preprint arXiv:2304.02643}} (\bibinfo{year}{2023}).
\newblock


\bibitem[Krausz et~al\mbox{.}(2022)]%
        {krausz2022comparison}
\bibfield{author}{\bibinfo{person}{Nikol Krausz}, \bibinfo{person}{Vivien Pot{\'o}}, \bibinfo{person}{J{\'a}nos~M{\'a}t{\'e} L{\'o}g{\'o}}, {and} \bibinfo{person}{{\'A}rp{\'a}d Barsi}.} \bibinfo{year}{2022}\natexlab{}.
\newblock \showarticletitle{Comparison of complex traffic junction descriptions in automotive standard formats}.
\newblock \bibinfo{journal}{\emph{Periodica Polytechnica Civil Engineering}} \bibinfo{volume}{66}, \bibinfo{number}{1} (\bibinfo{year}{2022}), \bibinfo{pages}{282--290}.
\newblock


\bibitem[Li et~al\mbox{.}(2022b)]%
        {li2022hdmapnet}
\bibfield{author}{\bibinfo{person}{Qi Li}, \bibinfo{person}{Yue Wang}, \bibinfo{person}{Yilun Wang}, {and} \bibinfo{person}{Hang Zhao}.} \bibinfo{year}{2022}\natexlab{b}.
\newblock \showarticletitle{Hdmapnet: An online hd map construction and evaluation framework}. In \bibinfo{booktitle}{\emph{2022 International Conference on Robotics and Automation (ICRA)}}. IEEE, \bibinfo{pages}{4628--4634}.
\newblock


\bibitem[Li et~al\mbox{.}(2022a)]%
        {li2022bevformer}
\bibfield{author}{\bibinfo{person}{Zhiqi Li}, \bibinfo{person}{Wenhai Wang}, \bibinfo{person}{Hongyang Li}, \bibinfo{person}{Enze Xie}, \bibinfo{person}{Chonghao Sima}, \bibinfo{person}{Tong Lu}, \bibinfo{person}{Yu Qiao}, {and} \bibinfo{person}{Jifeng Dai}.} \bibinfo{year}{2022}\natexlab{a}.
\newblock \showarticletitle{Bevformer: Learning bird’s-eye-view representation from multi-camera images via spatiotemporal transformers}. In \bibinfo{booktitle}{\emph{European conference on computer vision}}. Springer, \bibinfo{pages}{1--18}.
\newblock


\bibitem[Liao et~al\mbox{.}(2022)]%
        {liao2022maptr}
\bibfield{author}{\bibinfo{person}{Bencheng Liao}, \bibinfo{person}{Shaoyu Chen}, \bibinfo{person}{Xinggang Wang}, \bibinfo{person}{Tianheng Cheng}, \bibinfo{person}{Qian Zhang}, \bibinfo{person}{Wenyu Liu}, {and} \bibinfo{person}{Chang Huang}.} \bibinfo{year}{2022}\natexlab{}.
\newblock \showarticletitle{Maptr: Structured modeling and learning for online vectorized hd map construction}.
\newblock \bibinfo{journal}{\emph{arXiv preprint arXiv:2208.14437}} (\bibinfo{year}{2022}).
\newblock


\bibitem[Liao et~al\mbox{.}(2023)]%
        {liao2023maptrv2}
\bibfield{author}{\bibinfo{person}{Bencheng Liao}, \bibinfo{person}{Shaoyu Chen}, \bibinfo{person}{Yunchi Zhang}, \bibinfo{person}{Bo Jiang}, \bibinfo{person}{Qian Zhang}, \bibinfo{person}{Wenyu Liu}, \bibinfo{person}{Chang Huang}, {and} \bibinfo{person}{Xinggang Wang}.} \bibinfo{year}{2023}\natexlab{}.
\newblock \showarticletitle{Maptrv2: An end-to-end framework for online vectorized hd map construction}.
\newblock \bibinfo{journal}{\emph{arXiv preprint arXiv:2308.05736}} (\bibinfo{year}{2023}).
\newblock


\bibitem[Liu et~al\mbox{.}(2023a)]%
        {liu2023visual}
\bibfield{author}{\bibinfo{person}{Haotian Liu}, \bibinfo{person}{Chunyuan Li}, \bibinfo{person}{Qingyang Wu}, {and} \bibinfo{person}{Yong~Jae Lee}.} \bibinfo{year}{2023}\natexlab{a}.
\newblock \showarticletitle{Visual instruction tuning}.
\newblock \bibinfo{journal}{\emph{arXiv preprint arXiv:2304.08485}} (\bibinfo{year}{2023}).
\newblock


\bibitem[Liu et~al\mbox{.}(2021)]%
        {liu2021condlanenet}
\bibfield{author}{\bibinfo{person}{Lizhe Liu}, \bibinfo{person}{Xiaohao Chen}, \bibinfo{person}{Siyu Zhu}, {and} \bibinfo{person}{Ping Tan}.} \bibinfo{year}{2021}\natexlab{}.
\newblock \showarticletitle{Condlanenet: a top-to-down lane detection framework based on conditional convolution}. In \bibinfo{booktitle}{\emph{Proceedings of the IEEE/CVF International Conference on Computer Vision}}. \bibinfo{pages}{3773--3782}.
\newblock


\bibitem[Liu et~al\mbox{.}(2023b)]%
        {liu2023detection}
\bibfield{author}{\bibinfo{person}{Shilong Liu}, \bibinfo{person}{Tianhe Ren}, \bibinfo{person}{Jiayu Chen}, \bibinfo{person}{Zhaoyang Zeng}, \bibinfo{person}{Hao Zhang}, \bibinfo{person}{Feng Li}, \bibinfo{person}{Hongyang Li}, \bibinfo{person}{Jun Huang}, \bibinfo{person}{Hang Su}, \bibinfo{person}{Jun Zhu}, {et~al\mbox{.}}} \bibinfo{year}{2023}\natexlab{b}.
\newblock \showarticletitle{Detection Transformer with Stable Matching}.
\newblock \bibinfo{journal}{\emph{arXiv preprint arXiv:2304.04742}} (\bibinfo{year}{2023}).
\newblock


\bibitem[Liu et~al\mbox{.}(2023c)]%
        {liu2023vectormapnet}
\bibfield{author}{\bibinfo{person}{Yicheng Liu}, \bibinfo{person}{Tianyuan Yuan}, \bibinfo{person}{Yue Wang}, \bibinfo{person}{Yilun Wang}, {and} \bibinfo{person}{Hang Zhao}.} \bibinfo{year}{2023}\natexlab{c}.
\newblock \showarticletitle{Vectormapnet: End-to-end vectorized hd map learning}. In \bibinfo{booktitle}{\emph{International Conference on Machine Learning}}. PMLR, \bibinfo{pages}{22352--22369}.
\newblock


\bibitem[Loshchilov and Hutter(2017)]%
        {loshchilov2017adamw}
\bibfield{author}{\bibinfo{person}{Ilya Loshchilov} {and} \bibinfo{person}{Frank Hutter}.} \bibinfo{year}{2017}\natexlab{}.
\newblock \showarticletitle{Decoupled weight decay regularization}.
\newblock \bibinfo{journal}{\emph{arXiv preprint arXiv:1711.05101}} (\bibinfo{year}{2017}).
\newblock


\bibitem[Pan et~al\mbox{.}(2020)]%
        {pan2020vpn}
\bibfield{author}{\bibinfo{person}{Bowen Pan}, \bibinfo{person}{Jiankai Sun}, \bibinfo{person}{Ho~Yin~Tiga Leung}, \bibinfo{person}{Alex Andonian}, {and} \bibinfo{person}{Bolei Zhou}.} \bibinfo{year}{2020}\natexlab{}.
\newblock \showarticletitle{Cross-view semantic segmentation for sensing surroundings}.
\newblock \bibinfo{journal}{\emph{IEEE Robotics and Automation Letters}} \bibinfo{volume}{5}, \bibinfo{number}{3} (\bibinfo{year}{2020}), \bibinfo{pages}{4867--4873}.
\newblock


\bibitem[Peng et~al\mbox{.}(2023)]%
        {peng2023bevsegformer}
\bibfield{author}{\bibinfo{person}{Lang Peng}, \bibinfo{person}{Zhirong Chen}, \bibinfo{person}{Zhangjie Fu}, \bibinfo{person}{Pengpeng Liang}, {and} \bibinfo{person}{Erkang Cheng}.} \bibinfo{year}{2023}\natexlab{}.
\newblock \showarticletitle{BEVSegFormer: Bird's Eye View Semantic Segmentation From Arbitrary Camera Rigs}. In \bibinfo{booktitle}{\emph{Proceedings of the IEEE/CVF Winter Conference on Applications of Computer Vision}}. \bibinfo{pages}{5935--5943}.
\newblock


\bibitem[Qiao et~al\mbox{.}(2023a)]%
        {qiao2023bemapnet}
\bibfield{author}{\bibinfo{person}{Limeng Qiao}, \bibinfo{person}{Wenjie Ding}, \bibinfo{person}{Xi Qiu}, {and} \bibinfo{person}{Chi Zhang}.} \bibinfo{year}{2023}\natexlab{a}.
\newblock \showarticletitle{End-to-End Vectorized HD-Map Construction With Piecewise Bezier Curve}. In \bibinfo{booktitle}{\emph{Proceedings of the IEEE/CVF Conference on Computer Vision and Pattern Recognition}}. \bibinfo{pages}{13218--13228}.
\newblock


\bibitem[Qiao et~al\mbox{.}(2023b)]%
        {qiao2023machmap}
\bibfield{author}{\bibinfo{person}{Limeng Qiao}, \bibinfo{person}{Yongchao Zheng}, \bibinfo{person}{Peng Zhang}, \bibinfo{person}{Wenjie Ding}, \bibinfo{person}{Xi Qiu}, \bibinfo{person}{Xing Wei}, {and} \bibinfo{person}{Chi Zhang}.} \bibinfo{year}{2023}\natexlab{b}.
\newblock \showarticletitle{MachMap: End-to-End Vectorized Solution for Compact HD-Map Construction}.
\newblock \bibinfo{journal}{\emph{arXiv preprint arXiv:2306.10301}} (\bibinfo{year}{2023}).
\newblock


\bibitem[Release.(2019)]%
        {NDS}
\bibfield{author}{\bibinfo{person}{NDS Open Lane Model~1.0 Release.}} \bibinfo{year}{2019}\natexlab{}.
\newblock \showarticletitle{http://www.openlanemodel.org/}.
\newblock


\bibitem[Tabelini et~al\mbox{.}(2021)]%
        {tabelini2021laneatt}
\bibfield{author}{\bibinfo{person}{Lucas Tabelini}, \bibinfo{person}{Rodrigo Berriel}, \bibinfo{person}{Thiago~M Paixao}, \bibinfo{person}{Claudine Badue}, \bibinfo{person}{Alberto~F De~Souza}, {and} \bibinfo{person}{Thiago Oliveira-Santos}.} \bibinfo{year}{2021}\natexlab{}.
\newblock \showarticletitle{Keep your eyes on the lane: Real-time attention-guided lane detection}. In \bibinfo{booktitle}{\emph{Proceedings of the IEEE/CVF conference on computer vision and pattern recognition}}. \bibinfo{pages}{294--302}.
\newblock


\bibitem[Tao et~al\mbox{.}(2020a)]%
        {tao2020hierarchical}
\bibfield{author}{\bibinfo{person}{Andrew Tao}, \bibinfo{person}{Karan Sapra}, {and} \bibinfo{person}{Bryan Catanzaro}.} \bibinfo{year}{2020}\natexlab{a}.
\newblock \showarticletitle{Hierarchical multi-scale attention for semantic segmentation}.
\newblock \bibinfo{journal}{\emph{arXiv preprint arXiv:2005.10821}} (\bibinfo{year}{2020}).
\newblock


\bibitem[Tao et~al\mbox{.}(2020b)]%
        {tao2020hmsa}
\bibfield{author}{\bibinfo{person}{Andrew Tao}, \bibinfo{person}{Karan Sapra}, {and} \bibinfo{person}{Bryan Catanzaro}.} \bibinfo{year}{2020}\natexlab{b}.
\newblock \showarticletitle{Hierarchical multi-scale attention for semantic segmentation}.
\newblock \bibinfo{journal}{\emph{arXiv preprint arXiv:2005.10821}} (\bibinfo{year}{2020}).
\newblock


\bibitem[Wang et~al\mbox{.}(2022)]%
        {wang2022ganet}
\bibfield{author}{\bibinfo{person}{Jinsheng Wang}, \bibinfo{person}{Yinchao Ma}, \bibinfo{person}{Shaofei Huang}, \bibinfo{person}{Tianrui Hui}, \bibinfo{person}{Fei Wang}, \bibinfo{person}{Chen Qian}, {and} \bibinfo{person}{Tianzhu Zhang}.} \bibinfo{year}{2022}\natexlab{}.
\newblock \showarticletitle{A keypoint-based global association network for lane detection}. In \bibinfo{booktitle}{\emph{Proceedings of the IEEE/CVF Conference on Computer Vision and Pattern Recognition}}. \bibinfo{pages}{1392--1401}.
\newblock


\bibitem[Wang et~al\mbox{.}(2020)]%
        {wang2020hrnet}
\bibfield{author}{\bibinfo{person}{Jingdong Wang}, \bibinfo{person}{Ke Sun}, \bibinfo{person}{Tianheng Cheng}, \bibinfo{person}{Borui Jiang}, \bibinfo{person}{Chaorui Deng}, \bibinfo{person}{Yang Zhao}, \bibinfo{person}{Dong Liu}, \bibinfo{person}{Yadong Mu}, \bibinfo{person}{Mingkui Tan}, \bibinfo{person}{Xinggang Wang}, {et~al\mbox{.}}} \bibinfo{year}{2020}\natexlab{}.
\newblock \showarticletitle{Deep high-resolution representation learning for visual recognition}.
\newblock \bibinfo{journal}{\emph{IEEE transactions on pattern analysis and machine intelligence}} \bibinfo{volume}{43}, \bibinfo{number}{10} (\bibinfo{year}{2020}), \bibinfo{pages}{3349--3364}.
\newblock


\bibitem[Wu et~al\mbox{.}(2023)]%
        {wu20231st}
\bibfield{author}{\bibinfo{person}{Dongming Wu}, \bibinfo{person}{Fan Jia}, \bibinfo{person}{Jiahao Chang}, \bibinfo{person}{Zhuoling Li}, \bibinfo{person}{Jianjian Sun}, \bibinfo{person}{Chunrui Han}, \bibinfo{person}{Shuailin Li}, \bibinfo{person}{Yingfei Liu}, \bibinfo{person}{Zheng Ge}, {and} \bibinfo{person}{Tiancai Wang}.} \bibinfo{year}{2023}\natexlab{}.
\newblock \showarticletitle{The 1st-place Solution for CVPR 2023 OpenLane Topology in Autonomous Driving Challenge}.
\newblock \bibinfo{journal}{\emph{arXiv preprint arXiv:2306.09590}} (\bibinfo{year}{2023}).
\newblock


\bibitem[Xia et~al\mbox{.}(2022a)]%
        {duarus2022}
\bibfield{author}{\bibinfo{person}{Deguo Xia}, \bibinfo{person}{Jizhou Huang}, \bibinfo{person}{Jianzhong Yang}, \bibinfo{person}{Xiyan Liu}, {and} \bibinfo{person}{Haifeng Wang}.} \bibinfo{year}{2022}\natexlab{a}.
\newblock \showarticletitle{DuARUS: Automatic Geo-object Change Detection with Street View Imagery for Updating Road Database at Baidu Maps}. In \bibinfo{booktitle}{\emph{Proceedings of the 31st ACM International Conference on Information and Knowledge Management}}.
\newblock


\bibitem[Xia et~al\mbox{.}(2022b)]%
        {dutraffic2022}
\bibfield{author}{\bibinfo{person}{Deguo Xia}, \bibinfo{person}{Xiyan Liu}, \bibinfo{person}{Wei Zhang}, \bibinfo{person}{Hui Zhao}, \bibinfo{person}{Chengzhou Li}, \bibinfo{person}{Weiming Zhang}, \bibinfo{person}{Jizhou Huang}, {and} \bibinfo{person}{Haifeng Wang}.} \bibinfo{year}{2022}\natexlab{b}.
\newblock \showarticletitle{DuTraffic: Live Traffic Condition Prediction with Trajectory Data and Street Views at Baidu Maps}. In \bibinfo{booktitle}{\emph{Proceedings of the 31st ACM International Conference on Information and Knowledge Management}}.
\newblock


\bibitem[Yang et~al\mbox{.}(2022)]%
        {duare2022}
\bibfield{author}{\bibinfo{person}{Jianzhong Yang}, \bibinfo{person}{Xiaoqing Ye}, \bibinfo{person}{Bin Wu}, \bibinfo{person}{Yanlei Gu}, \bibinfo{person}{Ziyu Wang}, \bibinfo{person}{Deguo Xia}, {and} \bibinfo{person}{Jizhou Huang}.} \bibinfo{year}{2022}\natexlab{}.
\newblock \showarticletitle{DuARE: Automatic Road Extraction with Aerial Images and Trajectory Data at Baidu Maps}. In \bibinfo{booktitle}{\emph{Proceedings of the 28th ACM SIGKDD Conference on Knowledge Discovery and Data Mining}}. \bibinfo{pages}{4321--4331}.
\newblock


\bibitem[Zhang et~al\mbox{.}(2023a)]%
        {zhang2023mpvr}
\bibfield{author}{\bibinfo{person}{Gongjie Zhang}, \bibinfo{person}{Jiahao Lin}, \bibinfo{person}{Shuang Wu}, \bibinfo{person}{Yilin Song}, \bibinfo{person}{Zhipeng Luo}, \bibinfo{person}{Yang Xue}, \bibinfo{person}{Shijian Lu}, {and} \bibinfo{person}{Zuoguan Wang}.} \bibinfo{year}{2023}\natexlab{a}.
\newblock \showarticletitle{Online Map Vectorization for Autonomous Driving: A Rasterization Perspective}.
\newblock \bibinfo{journal}{\emph{arXiv preprint arXiv:2306.10502}} (\bibinfo{year}{2023}).
\newblock


\bibitem[Zhang et~al\mbox{.}(2023b)]%
        {zhang2023gemap}
\bibfield{author}{\bibinfo{person}{Zhixin Zhang}, \bibinfo{person}{Yiyuan Zhang}, \bibinfo{person}{Xiaohan Ding}, \bibinfo{person}{Fusheng Jin}, {and} \bibinfo{person}{Xiangyu Yue}.} \bibinfo{year}{2023}\natexlab{b}.
\newblock \showarticletitle{Online Vectorized HD Map Construction using Geometry}.
\newblock \bibinfo{journal}{\emph{arXiv preprint arXiv:2312.03341}} (\bibinfo{year}{2023}).
\newblock


\end{thebibliography}

\end{document}